\documentclass{ifacconf}

\usepackage{verbatim}
\usepackage{stackengine}
\usepackage{natbib, caption}
\usepackage{epsfig, enumerate}
\usepackage{array, multirow, dsfont}
\usepackage{subeqnarray, mathtools}
\usepackage{booktabs, threeparttable}
\usepackage{graphics, graphicx, subfigure,color}
\usepackage{mathtools, amsmath, amssymb, mathrsfs}
\usepackage[noend, ruled, vlined, algo2e]{algorithm2e}
\SetKwInput{KwInit}{Initialize}

\newtheorem{assumption}{Assumption}

\newcommand{\dist}{\text{dist}}
\newcommand{\sgn}{\text{sgn}}

\begin{document}
\begin{frontmatter}        

\title{\large Reactive Planning based Control for Mobile Robots in Obstacle-Cluttered Environments \thanksref{footnoteinfo}}
\thanks[footnoteinfo]{This work was supported by the National Natural Science Foundation of China under Grants 62573085, 62273320 and 08120003, the Fundamental Research Funds for the Central Universities under Grant DUT22RT(3)090, the LiaoNing Revitalization Talents Program (XLYC2403048) and the robotic AI-Scientist platform of Chinese Academy of Sciences.}

\author[1]{Li Tan},
\author[1]{Junlin Xiong},
\author[3]{Yan Wang},
\author[2]{Wei Ren}

\address[1]{Department of Automation, University of Science and Technology of China, Hefei, 230031, China. (emails: tlllll9@mail.ustc.edu.cn, xiong77@ustc.edu.cn).}
\address[3]{School of Intelligence Science and Engineering, Harbin Institute of Technology Shenzhen, Shenzhen 515100, China (email: wang.yan@hit.edu.cn)}
\address[2]{School of Control Science and Engineering, Dalian University of Technology, Dalian 116024, China. (email: wei.ren@dlut.edu.cn)}

\begin{abstract}
	This paper addresses the motion control problem for mobile robots in obstacle-cluttered environments. The mobile robot has partial environment information only, and aims to move from an initial position to a target position without collisions. For this purpose, a reactive planning based control strategy (RPCS) is proposed. First, the initial and target positions are connected as a reference trajectory. Then, a reactive planning strategy (RPS) is developed to ensure the collision avoidance by modifying the reference trajectory locally based on the partial environment information. Next, an adaptive tracking control strategy (ATCS) is proposed to track the reference trajectory with potentially local modifications via the discretization techniques. Finally, the RPS and ATCS are combined to establish the RPCS, whose efficacy and advantages are illustrated by numerical examples.
\end{abstract}


\end{frontmatter}

\section{Introduction}
\label{sec-intro}

Trajectory generation is essential for mobile robots to perform various missions, such as transportation, search and surveillance \citep{Kagan2019autonomous}. Due to the existence of obstacles, collision avoidance is a challenging issue in the trajectory generation. For the fully-known environment, there exist optimization-based \citep{Chen2015iSCP}, sampling-based \citep{Sertac2011Sampling-based} and search-based \citep{Sakcak2020A*} global approaches to ensure the collision avoidance. However, the robot sensing radii are limited, and only partial environment is known, which results in the difficulties in the collision avoidance. To this end, local approaches are developed and can be classified into the two directions. The first direction is to implement the global approaches, once the known environment information changes \citep{Fox1997DWA, Jing2016Online, Jankovic2018RobustCBF, Chen2019CILQR, Vrushabh20233DGSSE}. The repeated implementation ensures the collision avoidance, while it causes repetitive computation and is time-consuming in obstacle-cluttered environments.

To deal with this issue, the second direction composed of two steps is developed. The first step is to generate reference trajectories based on the partially-known environment information. The second step is to track the reference trajectories while modifying the reference trajectories for the collision avoidance. The tracking controller is designed locally, which avoids the repetitive computation and results in the advantage in terms of time-efficiency \citep{Liu2017Planning, Zhong2020HPP}. The existing local modifications are implemented by three ways: (i) imposing additional positional constraints \citep{Liu2017Planning, Park2022Online, Li2024Geometry}, (ii) choosing temporary target positions \citep{Helen2018Safe, Xue2024Combining}, (iii) selecting feasible motion directions \citep{Sun2017CushionRobot, Zhong2020HPP,  Zhao2022Scalable, Liu2023OptimalDirection}. The first way is valid through the geometric calculation, which results in huge computation complexity in the cluttered environments \citep{Liu2017Planning, Park2022Online, Li2024Geometry}. In the second way, temporary target positions are chosen via sampling techniques \citep{Helen2018Safe} or recurrent neural networks \citep{Xue2024Combining}, which require enough samples/data to ensure the feasibility. The third way is based on the geometric analysis of the obstacles and results in abrupt jumps \citep{Sun2017CushionRobot, Zhao2022Scalable, Liu2023OptimalDirection}, which are not practical due to the kinematic constraints. To solve this issue, the third way is adjusted via an optimization problem in \citep{Zhong2020HPP}, while the search algorithm is involved and results in large time and memory requirement, especially for large environments. As a result, modifying the reference trajectories smoothly with less time and memory requirement is still challenging.

In this paper, we investigate the motion control problem for mobile robots with partial information of the obstacle-cluttered environments. The robot task is to move from an initial position to a target position while ensuring the collision avoidance. To accomplish the task, we propose a novel local method, which is called a \emph{reactive planning based control strategy (RPCS)}. To begin with, a reference trajectory is generated by connecting the initial and target positions with a line segment. Next, to ensure the collision avoidance, cubic-polynomial techniques are used to develop a reactive planning strategy (RPS), which avoids both searching and sampling and is our first contribution. Specifically, the necessity of modifying the reference trajectory is checked first, and then cubic polynomials are used to modify the reference trajectory locally if necessary. With the move of the robot, the known environment information is updated, and thus the reference trajectory may be locally modified again afterwards. To track the reference trajectory with potentially local modifications, discretization techniques are adopted to develop an adaptive tracking control strategy (ATCS), which is our second contribution. Combining the RPS with the ATCS, the RPCS is established to enable the robot to accomplish its task via a smooth and globally collision-free trajectory. Compared to the existing local approaches, the RPCS ensures the trajectory smoothness and the motion direction continuity through the cubic-polynomial techniques in the RPS and the connection conditions in the ATCS. In addition, the RPCS has advantages in terms of the computation time and memory requirement due to the independence of the search/sampling-based algorithms, and is applicable to the mobile robots with diverse dynamics.

\textbf{Organization:} Section \ref{sec-problem} formulates the problem to be studied. The RPS and ATCS are developed in Sections \ref{sec-RPS}-\ref{sec-ATCS} to establish the RPCS. The simulation results are provided in Section \ref{sec-simulation}, followed by the conclusion in Section \ref{sec-conclusion}. 

\textbf{Notation:} Let $\mathbb{N}_{+}:=\{1,2,\ldots\}, \mathbb{R}:=(-\infty, +\infty)$ and $\mathbb{R}^{n}$ be the $n$-dimensional Euclidean space. $\|\cdot\|$ is the Euclidean norm in $\mathbb{R}^{n}$. $|\cdot|$ is the cardinality of a set. Given $x\in\mathbb{R}^{n}$ and $y\in\mathbb{R}^{m}$, $(x, y):=[x^{\top}~y^{\top}]^{\top}$. The function $\sgn:\mathbb{R}\rightarrow\{1,-1\}$ is defined as follows: for any $x\in\mathbb{R}$, $\sgn(x)=1$, if $x>0$; $\sgn(x)=-1$, if $x\le0$. For $p\in\mathbb{R}^{n}$ and $\mathcal{A},\mathcal{B}\subset\mathbb{R}^{n}$, $\dist(p,\mathcal{A})=\min_{x\in\mathcal{A}}\|p-x\|$ and $\mathcal{A}\setminus\mathcal{B}:=\{x: x\in\mathcal{A},x\notin\mathcal{B}\}$. 

\section{Problem Formulation}
\label{sec-problem}

Consider a mobile robot with the following dynamics:
\begin{align}
	\label{eqn-1}
	\dot{\xi}(t)=f(\xi(t),u(t)),
\end{align}
where $\xi(t):=(p(t),\zeta(t))\in\mathbb{R}^{n}$ is the state with the position state $p(t)\in\mathbb{R}^{2}$ and the non-position state $\zeta(t)\in\mathbb{R}^{n-2}$, and $u(t)\in\mathbb{U}$ is the control input with a compact set $\mathbb{U}\subset\mathbb{R}^{m}$. The function $f:\mathbb{R}^{n}\times\mathbb{R}^{m}\rightarrow\mathbb{R}^{n}$ is locally Lipschitz, which means that given any initial state, there exists a unique solution to \eqref{eqn-1} \cite[Thm.~3.1]{2001Nonlinear}. 

The robot is in a 2D environment with $K\in\mathbb{N}_{+}$ static obstacles. The regions occupied by the robot and the obstacles are convex and are respectively denoted as $\mathscr{R}(t),\mathscr{O}_{k}\subset\mathbb{R}^{2}$, where $k\in\mathbb{K}:=\{1,\ldots,K\}$. The size of the robot is determined by the two points with the largest distance in the occupied region $\mathscr{R}(t)$, and thus is defined as
\begin{align*}
	r:=\max\{\|p_{1}-p_{2}\|:p_{1},p_{2}\in\mathscr{R}(t)\}.
\end{align*}
Similarly, the size of obstacle $k$ is defined as
\begin{align*}
	r_{k}:=\max\{\|p_{1}-p_{2}\|:p_{1},p_{2}\in\mathscr{O}_{k}\}.
\end{align*}
The sensing region of the robot is a disk
\begin{align*}
	\mathscr{S}(t):=\{\mathfrak{p}\in\mathbb{R}^{2}:\|\mathfrak{p}-p(t)\|\le\gamma\},
\end{align*}
where $\gamma>0$ is the sensing radius. In order to enable the robot to avoid the obstacle collisions, the following assumptions are made.

\begin{assumption} 
\label{ass-1}
$\gamma>r+\max_{k\in\mathbb{K}} r_{k}$.
\end{assumption}

\begin{assumption} 
\label{ass-2}
$\mathscr{O}_{k}$ will be always known by the robot after the time $t>0$, if $\mathscr{O}_{k}\subset\mathscr{S}(t)$.
\end{assumption}

In Assumption \ref{ass-1}, the gap between $\gamma$ and $r+\max_{k\in\mathbb{K}} r_{k}$ is to enable the robot to avoid the collisions. From Assumption \ref{ass-2}, the index set of the obstacles known by the robot is 
\begin{align*}
    \mathscr{K}(t):=\{k\in\mathbb{K}:\exists t^{\prime}\le t\text{ s.t. } \mathscr{O}_{k}\subset\mathscr{S}(t^{\prime})\}.
\end{align*}
With the move of the robot, $|\mathscr{K}(t)|\in[0,K]$ is non-decreasing. That is, the robot only has partial environment information $\bigcup_{k\in\mathscr{K}(t)}\mathscr{O}_{k}$, which will not decrease. Under Assumptions \ref{ass-1}-\ref{ass-2}, we aim to propose a control strategy such that the robot can accomplish the task of moving from an initial position $s:=(x_{s},y_{s})\in\mathbb{R}^{2}$ to a target position $g:=(x_{g},y_{g})\in\mathbb{R}^{2}$ while ensuring the collision avoidance. To this end, an RPCS composed of an RPS and an ATCS is developed.

\begin{figure}[!t] 
	\begin{center}
		\begin{picture}(180, 17)
			\put(50,-9){\resizebox{32mm}{10mm}{\includegraphics[width=2.5in]{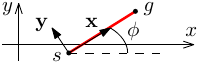}}}
		\end{picture}
	\end{center}
	\caption{Illustration of the local coordinate frame.}
	\label{fig-1}
\end{figure}

\section{Reactive Planning}
\label{sec-RPS}

In this section, the RPS is proposed to generate a locally collision-free reference trajectory. To simplify the calculation afterwards, we construct a local coordinate frame below: the origin is $s$, the $\mathbf{x}$-axis is the line connecting $s$ and $g$ with the direction from $s$ to $g$, and the $\mathbf{y}$-axis is obtained via rotating the $\mathbf{x}$-axis counterclockwise by $0.5\pi$; see Fig.~\ref{fig-1}. The angle between the $x$- and $\mathbf{x}$-axes is 
\begin{align*}
	\phi=\left\{\begin{aligned}
		&\arctan(y_{g}-y_{s})/(x_{g}-x_{s}), 
		&&\text{if } x_{g}\ne x_{s},  \\
		&0.5\pi\sgn(y_{g}-y_{s}),
		&&\text{if } x_{g}= x_{s}.
	\end{aligned}\right.
\end{align*}
The global and local coordinate frames can be transformed mutually via the rotation and translation techniques \cite[Ch.~12]{MechanicswithMathematica2018}. That is,
\begin{align}
	\label{eqn-2}
	\mathbf{p}=\mathbf{R}\left(p-s\right),\quad p=\mathbf{R}^{-1}\mathbf{p}+s,
\end{align}
where $p,\mathbf{p}\in\mathbb{R}^{2}$ are the positions in the global and local coordinate frames and $\mathbf{R}:=\begin{bmatrix}\begin{smallmatrix}
\cos\phi & \sin\phi \\ -\sin\phi & \cos\phi
\end{smallmatrix}\end{bmatrix}\in\mathbb{R}^{2\times2}$ is the transformation matrix. From \eqref{eqn-2}, the initial and target positions in the local coordinate frame are
\begin{align*}
	\mathbf{s}:=(\mathbf{x}_{s},\mathbf{y}_{s})=(0,0), \quad \mathbf{g}:=(\mathbf{x}_{g},\mathbf{y}_{g})=(\|g-s\|,0).
\end{align*}
Hence, the reference trajectory can be represented as
\begin{align}
	\label{eqn-3}
	\mathcal{L}(t):=\left\{\mathbf{p}:=(\mathbf{x},\mathbf{y})\in\mathbb{R}^{2}:\mathbf{x}\in\mathcal{X},\mathbf{y}=\ell(\mathbf{x},t)\right\},
\end{align}
where $\mathcal{X}:=[0,\mathbf{x}_{g}]$ and $\ell:[0,\mathbf{x}_{g}]\times[0,+\infty)\rightarrow\mathbb{R}$ is a continuously differentiable function. At the initial time $t=0$, the reference trajectory $\mathcal{L}_{0}:=\mathcal{L}(0)$ is the line segment connecting $\mathbf{s}$ and $\mathbf{g}$, which means that $\ell(\mathbf{x},0)=0$. In order to avoid the obstacle collisions, the reference trajectory is modified locally via updating $\ell(\mathbf{x},t)$. 

In order to check the modification necessity, the occupied region of each obstacle $k\in\mathscr{K}(t)$ is enlarged with the robot size $r$. The resulting enlarged occupied region in the local coordinate frame is denoted as $\mathcal{E}_{k}\subset\mathbb{R}^{2},k\in\mathscr{K}(t)$. With $\bigcup_{k\in\mathscr{K}(t)}\mathcal{E}_{k}$, we can determine the obstacles blocking the reference trajectory $\mathcal{L}(t)$ to be tracked, whose index set is
\begin{align}
	\label{eqn-4}
	\mathcal{K}(t):=\{k\in\mathscr{K}(t):\mathcal{L}(t)\cap\mathcal{E}_{k}\ne\emptyset\}.
\end{align} 
If $\mathcal{K}(t)=\emptyset$, then $\mathcal{L}(t)$ is not blocked and thus does not need to be modified; otherwise, $\mathcal{L}(t)$ is blocked by at least one obstacle and needs to be modified. In the case $\mathcal{K}(t)\ne\emptyset$, it is not easy to avoid multiple obstacles simultaneously, and thus the involved obstacles will be bypassed in order. Since $\dist(\mathbf{p}(t),\mathcal{E}_{k})=\dist(\mathbf{p}(t),\mathcal{E}_{j})$ may hold for $k,j\in\mathcal{K}(t)$ and $k\ne j$, the bypassing order is determined by
\begin{align}
	\label{eqn-5}
    \min\{k:\arg\min\nolimits_{k\in\mathcal{K}(t)} \dist(\mathbf{p}(t),\mathcal{E}_{k})\},
\end{align}
where $\mathbf{p}(t):=(\mathbf{x}(t),\mathbf{y}(t))\in\mathbb{R}^{2}$ is the current position of the robot. From \eqref{eqn-5}, only one solution $k\in\mathcal{K}(t)$ is derived, which means that obstacle $k$ is one of the obstacles closest to the robot and will be bypassed first.

Once the local modification is necessary to avoid obstacle $k$, a turning trajectory $\mathcal{H}_{k}\subset\mathbb{R}^{2}$ is generated to modify $\mathcal{L}(t)$ locally. For this purpose, we first determine the turning direction, then generate the turning trajectory $\mathcal{H}_{k}$, and finally establish the modified reference trajectory.

\begin{figure}[!t] 
	\begin{center}
		\begin{picture}(230, 45)
			\put(-5,-7){\resizebox{85mm}{18mm}{\includegraphics[width=2.5in]{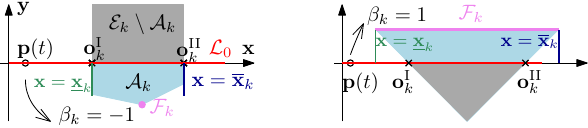}}}
		\end{picture}
	\end{center}
	\caption{Illustration of the case $(\mathbf{o}_{k}^{\text{I}}\in\mathcal{L}_{0})\wedge(\mathbf{o}_{k}^{\text{II}}\in\mathcal{L}_{0})$.}
	\label{fig-2}
\end{figure}

\subsubsection{\textbf{Turning Direction Determination.}}
Let $\beta_{k}=\{-1,1\}$ be the turning direction, where $\beta_{k}=1$ is to turn left and $\beta_{k}=-1$ is to turn right. Since $\mathcal{L}(t)$ is blocked by obstacle $k$, the first and last potential collision positions are the two intersections between the boundary of $\mathcal{E}_{k}$ and $\mathcal{L}(t)$, and are derived below:
\begin{align}
\label{eqn-6}
\begin{aligned}
\mathbf{o}_{k}^{\text{I}}&:=\mathop{\arg\min}\nolimits_{\mathbf{p}\in\mathcal{E}_{k}\cap\mathcal{L}(t)}\mathbf{x},  \\ \mathbf{o}_{k}^{\text{II}}&:=\mathop{\arg\max}\nolimits_{\mathbf{p}\in\mathcal{E}_{k}\cap\mathcal{L}(t)}\mathbf{x}.
\end{aligned}
\end{align}
Here $\mathcal{L}(t)$ can be the initial one or the modified one. Each of $\mathbf{o}_{k}^{\text{I}}$ and $\mathbf{o}_{k}^{\text{II}}$ is on a part of $\mathcal{L}_{0}$ or the turning trajectory $\mathcal{H}_{j}\subset\mathbb{R}^{2}$, which is generated previously to bypass obstacle $j\in\mathscr{K}(t)\setminus\{k\}$. Based on the locations of $\mathbf{o}_{k}^{\text{I}}$ and $\mathbf{o}_{k}^{\text{II}}$, we discuss the determination of $\beta_{k}$ below.
\begin{itemize}
	\item $(\mathbf{o}_{k}^{\text{I}}\in\mathcal{L}_{0})\wedge(\mathbf{o}_{k}^{\text{II}}\in\mathcal{L}_{0})$: $\mathcal{L}_{0}$ divides $\mathcal{E}_{k}$ into two parts. We let the robot turn to bypass the obstacle part with the smaller size along the $\mathbf{y}$-axis such that the turning trajectory is as short as possible; see Fig.~\ref{fig-2}. That is, the turning direction is chosen as  
	\begin{align*}
		\beta_{k}=\left\{\begin{aligned}
			&1, &&\text{if } \|\max\nolimits_{\mathbf{p}\in\mathcal{E}_{k}}\mathbf{y}\|<\|\min\nolimits_{\mathbf{p}\in\mathcal{E}_{k}}\mathbf{y}\|, \\
			&-1, &&\text{if } \|\max\nolimits_{\mathbf{p}\in\mathcal{E}_{k}}\mathbf{y}\|\ge\|\min\nolimits_{\mathbf{p}\in\mathcal{E}_{k}}\mathbf{y}\|.
		\end{aligned}\right.
	\end{align*}
	
	\item $(\mathbf{o}_{k}^{\text{I}}\in\mathcal{H}_{j})\vee (\mathbf{o}_{k}^{\text{II}}\in\mathcal{H}_{j})$: In this case, we need to ensure bypassing both $\mathcal{E}_{j}$ and $\mathcal{E}_{k}$, and thus
	\begin{align*}
		\beta_{k}=\beta_{j},
	\end{align*}
	where $\beta_{j}\in\{-1,1\}$ is the turning direction for bypassing $\mathcal{E}_{j}$. Note that $\beta_{k}$ can be also determined as in the first case, which however may result in the issue where the robot needs to move through the narrow corridor between $\mathcal{E}_{j}$ and $\mathcal{E}_{k}$; see Fig.~\ref{fig-3}. This issue may cause a high computation complexity in the control design and even obstacle collision.
\end{itemize}
From the above discussion, the turning direction is 
\begin{align}
	\label{eqn-7}
	\beta_{k}=\left\{\begin{aligned}
		\sgn(\|\min\nolimits_{\mathbf{p}\in\mathcal{E}_{k}}&\mathbf{y}\|-\|\max\nolimits_{\mathbf{p}\in\mathcal{E}_{k}}\mathbf{y}\|), \\ 
		\qquad &\text{if } (\mathbf{o}_{k}^{\text{I}}\in\mathcal{L}_{0})\wedge(\mathbf{o}_{k}^{\text{II}}\in\mathcal{L}_{0}), \\
		\beta_{j}, \qquad\qquad\quad &\text{if } (\mathbf{o}_{k}^{\text{I}}\in\mathcal{H}_{j})\vee (\mathbf{o}_{k}^{\text{II}}\in\mathcal{H}_{j}).
    \end{aligned}\right.
\end{align}

\subsubsection{\textbf{Turning Trajectory Generation.}}
\begin{figure}[!t] 
	\begin{center}
		\begin{picture}(230, 55)
			\put(15,-7){\resizebox{68mm}{21mm}{\includegraphics[width=2.5in]{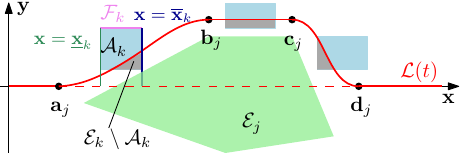}}}
		\end{picture}
	\end{center}
	\caption{Illustration of the case $(\mathbf{o}_{k}^{\text{I}}\in\mathcal{H}_{j})\vee (\mathbf{o}_{k}^{\text{II}}\in\mathcal{H}_{j})$.}
	\label{fig-3}
\end{figure}

With $\beta_{k}$, the obstacle part to be bypassed is determined as follows
\begin{align*}
	\mathcal{A}_{k}:=\{\mathbf{p}\in\mathcal{E}_{k}:\sgn(\mathbf{y}-\ell(\mathbf{x},t))=\beta_{k}\}.
\end{align*}
In $\mathcal{A}_{k}$, the points with the smallest abscissa are on the line $\mathbf{x}=\underline{\mathbf{x}}_{k}$, the points with the largest abscissa are on the line $\mathbf{x}=\overline{\mathbf{x}}_{k}$, and the points farthest from $\mathcal{L}(t)$ are in the set $\mathcal{F}_{k}:=\{\mathbf{p}\in\mathbb{R}^{2}:\mathbf{x}\in[\mathbf{x}_{k}^{\text{n}},\mathbf{x}_{k}^{\text{f}}],\mathbf{y}=\mathbf{y}_{k}\}$, where
\begin{align}
	\label{eqn-8}
	\begin{aligned}
		\underline{\mathbf{x}}_{k}&:=\min\nolimits_{\mathbf{p}\in\mathcal{A}_{k}}\mathbf{x}, \quad  \overline{\mathbf{x}}_{k}:=\max\nolimits_{\mathbf{p}\in\mathcal{A}_{k}}\mathbf{x}, \\ 
		\mathbf{x}_{k}^{\text{n}}&:=\mathop{\arg\min}\nolimits_{\mathbf{x}\in\mathbb{R}}\mathop{\arg\max}\nolimits_{\mathbf{p}\in\mathcal{A}_{k}}\|\mathbf{y}\|, \\
		\mathbf{x}_{k}^{\text{f}}&:=\mathop{\arg\max}\nolimits_{\mathbf{x}\in\mathbb{R}}\mathop{\arg\max}\nolimits_{\mathbf{p}\in\mathcal{A}_{k}}\|\mathbf{y}\|, \\
		\mathbf{y}_{k}&:=\beta_{k}(\max\nolimits_{\mathbf{p}\in\mathcal{A}_{k}}\|\mathbf{y}\|).
	\end{aligned}
\end{align}
Note that $\mathbf{x}_{k}^{\text{n}}=\mathbf{x}_{k}^{\text{f}}, \mathbf{x}_{k}^{\text{n}}=\underline{\mathbf{x}}_{k}$ or $\mathbf{x}_{k}^{\text{f}}=\overline{\mathbf{x}}_{k}$ may hold; see Figs.~\ref{fig-2}-\ref{fig-3}. In $\mathcal{F}_{k}$, the nearest and farthest points from the current position $\mathbf{p}(t)$ are $(\mathbf{x}_{k}^{\text{n}},\mathbf{y}_{k})$ and $(\mathbf{x}_{k}^{\text{f}},\mathbf{y}_{k})$, respectively. If $\mathcal{F}_{k}$ can be bypassed, then obstacle $k$ can be avoided. To this end, we generate the turning trajectory $\mathcal{H}_{k}$ with the following three parts.
\begin{itemize}
	\item The first part $\mathcal{H}_{k}^{\text{I}}\subset\mathbb{R}^{2}$ turns from the starting position $\mathbf{a}_{k}:=(\mathbf{x}_{k}^{\mathbf{a}},\mathbf{y}_{k}^{\mathbf{a}})\in\mathbb{R}^{2}$ to the ending position $\mathbf{b}_{k}:=(\mathbf{x}_{k}^{\mathbf{b}},\mathbf{y}_{k}^{\mathbf{b}})\in\mathbb{R}^{2}$ to prepare for bypassing $\mathcal{F}_{k}$. 
	
	\item The second part $\mathcal{H}_{k}^{\text{II}}\subset\mathbb{R}^{2}$ starts from $\mathbf{b}_{k}$ to the ending position $\mathbf{c}_{k}:=(\mathbf{x}_{k}^{\mathbf{c}},\mathbf{y}_{k}^{\mathbf{c}})\in\mathbb{R}^{2}$ to bypass $\mathcal{F}_{k}$. 
	
	\item The third part $\mathcal{H}_{k}^{\text{III}}\subset\mathbb{R}^{2}$ starts from $\mathbf{c}_{k}$ to the ending position $\mathbf{d}_{k}:=(\mathbf{x}_{k}^{\mathbf{d}},\mathbf{y}_{k}^{\mathbf{d}})\in\mathbb{R}^{2}$ to back to $\mathcal{L}(t)$.
\end{itemize}
Let $\eta_{k}^{\mathbf{a}},\eta_{k}^{\mathbf{b}},\eta_{k}^{\mathbf{c}},\eta_{k}^{\mathbf{d}}\in\mathbb{R}$ be the slopes at $\mathbf{a}_{k},\mathbf{b}_{k},\mathbf{c}_{k},\mathbf{d}_{k}$. The positions $\mathbf{a}_{k},\mathbf{b}_{k},\mathbf{c}_{k},\mathbf{d}_{k}$ and their slopes $\eta_{k}^{\mathbf{a}},\eta_{k}^{\mathbf{b}},\eta_{k}^{\mathbf{c}},\eta_{k}^{\mathbf{d}}$ are determined below to generate $\mathcal{H}_{k}$ via cubic polynomials \citep{Liu2022LaneChange}.

\begin{figure}[!t]
	\begin{center}
		\begin{picture}(170,60)
			\put(0,-10){\resizebox{58mm}{25mm}{\includegraphics[width=2.5in]{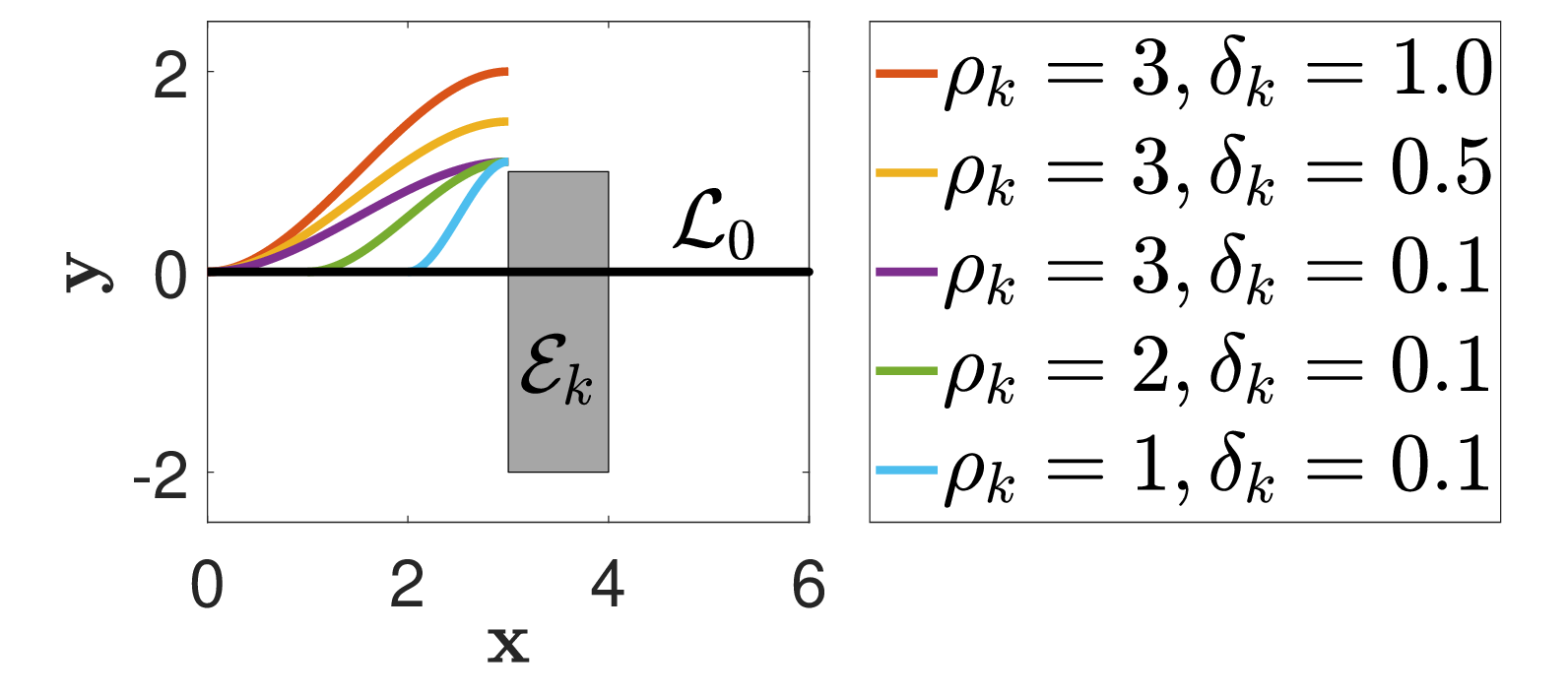}}}
		\end{picture}
	\end{center}
	\caption{Illustration of the effects of $\rho_{k}$ and $\delta_{k}$.}
	\label{fig-4}
\end{figure}

In order to choose $\mathbf{a}_{k}$, the position $\mathbf{p}(t)$ and the line $\mathbf{x}=\underline{\mathbf{x}}_{k}$ are considered. To bypass $\mathcal{A}_{k}$, the robot needs to start turning before the position $(\underline{\mathbf{x}}_{k}, \ell(\underline{\mathbf{x}}_{k},t))$, which can be triggered by the following condition
\begin{align}
    \label{eqn-9}
    \underline{\mathbf{x}}_{k}-\mathbf{x}(t)=\rho_{k},
\end{align}
where $\rho_{k}>0$ is a constant parameter, which will be determined later. Due to the limited sensing radius, \eqref{eqn-9} may be satisfied before the current time $t>0$. If $\mathbf{x}(t)<\underline{\mathbf{x}}_{k}$, then the following condition is added to enable the robot to move forward along the $\mathbf{x}$-axis. 
\begin{align*}
	\mathbf{x}_{k}^{\mathbf{a}}\ge\mathbf{x}(t).
\end{align*}
Otherwise, the obstacle avoidance takes priority over moving forward along the $\mathbf{x}$-axis, and thus only \eqref{eqn-9} is considered. In this way, either $\mathbf{x}_{k}^{\mathbf{a}}\le0$ or $\mathbf{x}_{k}^{\mathbf{a}}>0$, which is considered to choose $\mathbf{a}_{k}$ as
\begin{align}
	\label{eqn-10}
	\begin{aligned}
		\mathbf{x}_{k}^{\mathbf{a}}&=\left\{\begin{aligned}
			&\underline{\mathbf{x}}_{k}-\rho_{k}, &&\text{if } \mathbf{x}(t)\ge\underline{\mathbf{x}}_{k}, \\
			&\min\{\mathbf{x}(t),\underline{\mathbf{x}}_{k}-\rho_{k}\}, &&\text{if } \mathbf{x}(t)<\underline{\mathbf{x}}_{k}, \\
		\end{aligned}\right. \\  
		\mathbf{y}_{k}^{\mathbf{a}}&=\left\{\begin{aligned}
			&\mathbf{y}(t), &&\qquad\qquad\quad~\text{if } \mathbf{x}_{k}^{\mathbf{a}}<0, \\
			&\ell(\mathbf{x}_{k}^{\mathbf{a}},t), &&\qquad\qquad\quad~\text{if } \mathbf{x}_{k}^{\mathbf{a}}\ge0.
		\end{aligned}\right.
	\end{aligned}
\end{align}
To prepare for bypassing $\mathcal{F}_{k}$, the nearest position $(\mathbf{x}_{k}^{\text{n}},\mathbf{y}_{k})$ in $\mathcal{F}_{k}$ is utilized to choose $\mathbf{b}_{k}$ as 
\begin{align}
	\label{eqn-11}
	\mathbf{x}_{k}^{\mathbf{b}}=\mathbf{x}_{k}^{\text{n}},\quad \mathbf{y}_{k}^{\mathbf{b}}=\mathbf{y}_{k}+\beta_{k}\delta_{k},
\end{align}
where $\delta_{k}>0$ is constant and will be determined later. Connecting $\mathbf{a}_{k}$ and $\mathbf{b}_{k}$ via a cubic polynomial curve, we obtain the first part of the turning trajectory below:
\begin{align*}
	\mathcal{H}_{k}^{\text{I}}=\{\mathbf{p}\in\mathbb{R}^{2}:~ &\mathbf{x}\in[\mathbf{x}_{k}^{\mathbf{a}}, \mathbf{x}_{k}^{\mathbf{b}}], \\ &\mathbf{y}=\mathfrak{a}_{k}\mathbf{x}^{3}+\mathfrak{b}_{k}\mathbf{x}^{2}+\mathfrak{c}_{k}\mathbf{x}+\mathfrak{d}_{k}\},
\end{align*}
where $\mathfrak{a}_{k}, \mathfrak{b}_{k}, \mathfrak{c}_{k}, \mathfrak{d}_{k}\in\mathbb{R}$ are the coefficients derived from
\begin{align}
    \label{eqn-12}
    \begin{bmatrix}
        \mathbf{y}_{k}^{\mathbf{a}} \\ \mathbf{y}_{k}^{\mathbf{b}} \\ 	\eta_{k}^{\mathbf{a}} \\ \eta_{k}^{\mathbf{b}}
    \end{bmatrix}&=\begin{bmatrix}
        (\mathbf{x}_{k}^{\mathbf{a}})^{3} & (\mathbf{x}_{k}^{\mathbf{a}})^{2} & \mathbf{x}_{k}^{\mathbf{a}} & 1 \\
        (\mathbf{x}_{k}^{\mathbf{b}})^{3} & (\mathbf{x}_{k}^{\mathbf{b}})^{2} & \mathbf{x}_{k}^{\mathbf{b}} & 1 \\
        3(\mathbf{x}_{k}^{\mathbf{a}})^{2} & 2\mathbf{x}_{k}^{\mathbf{a}} & 1 & 0\\
        3(\mathbf{x}_{k}^{\mathbf{b}})^{2} & 2\mathbf{x}_{k}^{\mathbf{b}} & 1 & 0\\
    \end{bmatrix}
    \begin{bmatrix}
        \mathfrak{a}_{k} \\ \mathfrak{b}_{k} \\  \mathfrak{c}_{k} \\ \mathfrak{d}_{k}
    \end{bmatrix}
\end{align}
with the following slopes
\begin{align}
	\label{eqn-13}
	\eta_{k}^{\mathbf{a}}=\left\{\begin{aligned}
		&0, &&\text{if } \mathbf{x}_{k}^{\mathbf{a}}\le0, \\
		&\left.\tfrac{\partial\ell(\mathbf{x},t)}{\partial\mathbf{x}}\right|_{\mathbf{x}=\mathbf{x}^{\mathbf{a}}_{k}}, &&\text{if } \mathbf{x}_{k}^{\mathbf{a}}>0,
	\end{aligned}\right. \qquad \eta_{k}^{\mathbf{b}}=0,
\end{align}
which are to guarantee the smoothness of the turning trajectory. Note that the coefficients $\mathfrak{a}_{k},\mathfrak{b}_{k},\mathfrak{c}_{k},\mathfrak{d}_{k}$ derived from \eqref{eqn-12} are related the parameters $\rho_{k}$ and $\delta_{k}$. That is, both $\rho_{k}$ and $\delta_{k}$ have effects on $\mathcal{H}_{k}^{\text{I}}$. With the decrease of $\rho_{k}$ or the increase of $\delta_{k}$, $\mathcal{H}_{k}^{\text{I}}$ becomes longer and steeper; see Fig.~\ref{fig-4}. How to choose $\delta_{k}$ and $\rho_{k}$ appropriately depends on the considered system and the desired performance.

Similar to \eqref{eqn-11}, $(\mathbf{x}_{k}^{\text{f}},\mathbf{y}_{k})$ is used to set $\mathbf{c}_{k}$ as
\begin{align}
	\label{eqn-14}
	\mathbf{x}_{k}^{\mathbf{c}}=\mathbf{x}_{k}^{\text{f}},\quad \mathbf{y}_{k}^{\mathbf{c}}=\mathbf{y}_{k}+\beta_{k}\delta_{k}.
\end{align}
Since $\mathbf{y}_{k}^{\mathbf{b}}=\mathbf{y}_{k}^{\mathbf{c}}$, a line segment is adopted to connect $\mathbf{b}_{k}$ and $\mathbf{c}_{k}$ to generate the second part as follows
\begin{align*}
    \mathcal{H}_{k}^{\text{II}}=\{\mathbf{p}\in\mathbb{R}^{2}: \mathbf{x}\in[\mathbf{x}_{k}^{\mathbf{b}},\mathbf{x}_{k}^{\mathbf{c}}],\mathbf{y}=\mathbf{y}_{k}^{\mathbf{b}}\}.
\end{align*}
In this way, $\mathcal{H}_{k}^{\text{II}}$ is a point or a line segment. Note that $\mathcal{H}_{k}^{\text{II}}$ can be also generated via a cubic polynomial, which still results in the line segment connecting $\mathbf{b}_{k}$ and $\mathbf{c}_{k}$.

Similar to the choice of $\mathbf{a}_{k}$, the target position $\mathbf{g}$ and the line $\mathbf{x}=\overline{\mathbf{x}}_{k}$ are considered to select $\mathbf{d}_{k}$. In order to avoid the collision with obstacle $k$ during backing to $\mathcal{L}(t)$, we expect $\mathbf{x}_{k}^{\mathbf{d}}>\overline{\mathbf{x}}_{k}$ and thus trigger the return action by the following condition
\begin{align*}
	\mathbf{x}(t)-\overline{\mathbf{x}}_{k}=\rho_{k}.
\end{align*} 
From the above condition, $\mathbf{x}_{k}^{\mathbf{d}}\le\mathbf{x}_{g}$ or $\mathbf{x}_{k}^{\mathbf{d}}>\mathbf{x}_{g}$ may hold. Hence, $\mathbf{d}_{k}$ is set as
\begin{align}
    \label{eqn-15}
    \mathbf{x}_{k}^{\mathbf{d}}&=\overline{\mathbf{x}}_{k}+\rho_{k}, \qquad
    \mathbf{y}_{k}^{\mathbf{d}}=\left\{\begin{aligned}
    	&\ell(\mathbf{x}_{k}^{\mathbf{d}},t), &&\text{if } \mathbf{x}_{k}^{\mathbf{d}}\le\mathbf{x}_{g}, \\
    	&0, &&\text{if } \mathbf{x}_{k}^{\mathbf{d}}>\mathbf{x}_{g}.
    \end{aligned}\right.
\end{align}
To ensure the smoothness of the turning trajectory, we let 
\begin{align}
	\label{eqn-16}
	\eta_{k}^{\mathbf{c}}=0,\qquad \eta_{k}^{\mathbf{d}}=\left\{\begin{aligned}
		&\left.\tfrac{\partial\ell(\mathbf{x},t)}{\partial\mathbf{x}}\right|_{\mathbf{x}=\mathbf{x}^{\mathbf{d}}_{k}}, &&\text{if } \mathbf{x}_{k}^{\mathbf{d}}\le\mathbf{x}_{g}, \\
		&0, &&\text{if } \mathbf{x}_{k}^{\mathbf{d}}>\mathbf{x}_{g}.
	\end{aligned}\right.
\end{align}
With $\mathbf{c}_{k},\mathbf{d}_{k},\eta_{k}^{\mathbf{c}}$ and $\eta_{k}^{\mathbf{d}}$, we use a cubic polynomial function to generate the third part below:
\begin{align*}
	\mathcal{H}_{k}^{\text{III}}=\{\mathbf{p}\in\mathbb{R}^{2}:~ &\mathbf{x}\in[\mathbf{x}_{k}^{\mathbf{c}}, \mathbf{x}_{k}^{\mathbf{d}}], \\ &\mathbf{y}=\mathfrak{e}_{k}\mathbf{x}^{3}+\mathfrak{f}_{k}\mathbf{x}^{2}+\mathfrak{g}_{k}\mathbf{x}+\mathfrak{h}_{k}\},
\end{align*}
where $\mathfrak{e}_{k},\mathfrak{f}_{k}, \mathfrak{g}_{k},\mathfrak{h}_{k}\in\mathbb{R}$ are the coefficients, which are derived from the equation of the form \eqref{eqn-12} and are also related to the parameters $\rho_{k}$ and $\delta_{k}$.

Combining the three parts $\mathcal{H}_{k}^{\text{I}},\mathcal{H}_{k}^{\text{II}}$ and $\mathcal{H}_{k}^{\text{III}}$, we obtain the whole turning trajectory
\begin{align}
	\label{eqn-17}
	\mathcal{H}_{k}=\left\{\mathbf{p}\in\mathbb{R}^{2}:\mathbf{x}\in[\mathbf{x}_{k}^{\mathbf{a}},\mathbf{x}_{k}^{\mathbf{d}}],\mathbf{y}=h_{k}(\mathbf{x})\right\},
\end{align}
where
\begin{align*}
	h_{k}(\mathbf{x})=\left\{\begin{aligned}
		&\mathfrak{a}_{k}\mathbf{x}^{3}+\mathfrak{b}_{k}\mathbf{x}^{2}+\mathfrak{c}_{k}\mathbf{x}+\mathfrak{d}_{k},&&\text{if } \mathbf{x}\in[\mathbf{x}_{k}^{\mathbf{a}},\mathbf{x}_{k}^{\mathbf{b}}], \\
		&\mathbf{y}_{k}^{\mathbf{b}},&&\text{if } \mathbf{x}\in[\mathbf{x}_{k}^{\mathbf{b}},\mathbf{x}_{k}^{\mathbf{c}}], \\
		&\mathfrak{e}_{k}\mathbf{x}^{3}+\mathfrak{f}_{k}\mathbf{x}^{2}+\mathfrak{g}_{k}\mathbf{x}+\mathfrak{h}_{k},&&\text{if } \mathbf{x}\in[\mathbf{x}_{k}^{\mathbf{c}},\mathbf{x}_{k}^{\mathbf{d}}].
	\end{aligned}\right.
\end{align*}
For $\mathcal{H}_{k}$, it is not easy to tune the parameters $\rho_{k}$ and $\delta_{k}$ simultaneously. Hence, we fix $\{\rho_{k},\delta_{k}\}\setminus\{\bigstar_{k}\}$ and only choose $\bigstar_{k},\bigstar\in\{\rho,\delta\}$. In order to ensure the turning trajectory to be as short as possible, $\bigstar_{k}$ is derived from the following optimization problem:
\begin{align}
\label{eqn-18}
\begin{aligned}
\min\nolimits_{\bigstar_{k}>0} &~~ \bigstar_{k} \\
\text{subject to} &~~ \mathcal{H}_{k}\cap\mathcal{A}_{k}=\emptyset, ~~ \mathcal{H}_{k}\cap\mathcal{E}_{j}=\emptyset.
\end{aligned}
\end{align}
Note that the constraint $\mathcal{H}_{k}\cap\mathcal{E}_{j}=\emptyset$ is not needed in the case $(\mathbf{o}_{k}^{\text{I}}\in\mathcal{L}_{0})\wedge(\mathbf{o}_{k}^{\text{II}}\in\mathcal{L}_{0})$.

\subsubsection{\textbf{Modified Reference Trajectory Establishment.}}
From \eqref{eqn-10} and \eqref{eqn-15}, $\mathbf{a}_{k}\not\in\mathcal{L}(t)$ or $\mathbf{d}_{k}\not\in\mathcal{L}(t)$ may hold. Hence, $\mathcal{H}_{k}$ cannot be always inserted into $\mathcal{L}(t)$ to establish the modified reference trajectory directly. With $\mathcal{H}_{k}$, the valid parts of $\mathcal{L}(t)$ are
\begin{align*}
	\mathcal{L}_{k}^{1}:=\{\mathbf{p}\in\mathbb{R}^{2}:\mathbf{x}\in\mathcal{X}_{k}^{1},\mathbf{y}=\ell(\mathbf{x},t)\}, \\ \mathcal{L}_{k}^{2}:=\{\mathbf{p}\in\mathbb{R}^{2}:\mathbf{x}\in\mathcal{X}_{k}^{2},\mathbf{y}=\ell(\mathbf{x},t)\},
\end{align*}
where
\begin{align*}
\mathcal{X}_{k}^{1}=[0,\max\{\mathbf{x}(t),\mathbf{x}_{k}^{\mathbf{a}}\}], \quad 
\mathcal{X}_{k}^{2}=\left\{\begin{aligned}
&[\mathbf{x}_{k}^{\mathbf{d}},\mathbf{x}_{g}], &&\text{if } \mathbf{x}_{k}^{\mathbf{d}}<\mathbf{x}_{g}, \\
&\emptyset, &&\text{if } \mathbf{x}_{k}^{\mathbf{d}}\ge\mathbf{x}_{g}.
\end{aligned}\right.
\end{align*}
Since $\mathbf{g}\not\in\mathcal{L}_{k}^{2}$ may hold, the following two connection trajectories are used to connect $\mathcal{L}_{k}^{1},\mathcal{H}_{k},\mathcal{L}_{k}^{2}$ and $\mathbf{g}$ to establish the modified reference trajectory.
\begin{align*}
\mathcal{C}_{k}^{1}&:=\left\{\begin{aligned}
&\emptyset, &&\text{if } \mathbf{x}_{k}^{\mathbf{a}}\ge\mathbf{x}(t),\\
&\{\mathbf{p}\in\mathbb{R}^{2}:\mathbf{x}\in[\mathbf{x}_{k}^{\mathbf{a}},\mathbf{x}(t)],\mathbf{y}=\mathbf{y}(t)\},  &&\text{if } \mathbf{x}_{k}^{\mathbf{a}}<0,\\
&\{\mathbf{p}\in\mathbb{R}^{2}:\mathbf{x}\in[\mathbf{x}_{k}^{\mathbf{a}},\mathbf{x}(t)],\mathbf{y}=\ell(\mathbf{x},t)\}, &&\text{otherwise},
\end{aligned}\right. \\
\mathcal{C}_{k}^{2}&:=\left\{\begin{aligned}
&\emptyset, &&\qquad~~ \text{ if } \mathbf{x}_{k}^{\mathbf{d}}\le\mathbf{x}_{g}, \\
&\{\mathbf{p}\in\mathbb{R}^{2}:\mathbf{x}\in[\mathbf{x}_{g},\mathbf{x}_{k}^{\mathbf{d}}],\mathbf{y}=0\}, &&\qquad~~ \text{ if } \mathbf{x}_{k}^{\mathbf{d}}>\mathbf{x}_{g}.
\end{aligned}\right.
\end{align*}
Note that in the case $\mathbf{x}_{k}^{\mathbf{d}}\ne\mathbf{x}_{g}$, $\mathcal{C}_{k}^{2}=\emptyset$ and $\mathcal{L}_{k}^{2}=\emptyset$ will not hold simultaneously. When determining the turning direction in \eqref{eqn-7}, both $\mathcal{C}_{k}^{1}=\{\mathbf{p}:\mathbf{x}\in[\mathbf{x}_{k}^{\mathbf{a}},\mathbf{x}(t)],\mathbf{y}=\mathbf{y}(t)\}$ and $\mathcal{C}_{k}^{2}=\{\mathbf{p}:\mathbf{x}\in[\mathbf{x}_{g},\mathbf{x}_{k}^{\mathbf{d}}],\mathbf{y}=0\}$ are taken as the initial reference trajectory. Since no collision has occurred yet, $\mathcal{C}_{k}^{1}=\{\mathbf{p}:\mathbf{x}\in[\mathbf{x}_{k}^{\mathbf{a}},\mathbf{x}(t)],\mathbf{y}=\ell(\mathbf{x},t)\}\subset\mathcal{T}_{0}(t)$ is collision-free and thus will not be involved in the determination of the turning direction, where $\mathcal{T}_{0}(t):=\{\mathbf{p}\in\mathbb{R}^{2}:\mathbf{x}\in[0,\mathbf{x}(t)),\mathbf{y}=\ell(\mathbf{x},t)\}$ is the reference trajectory part that has already been tracked.

\begin{algorithm2e}[!t]
\SetAlgoLined
	\KwIn{$\mathcal{L}(t),\cup_{k\in\mathscr{K}(t)}\mathcal{E}_{k}$ }
	\While{$\mathcal{K}(t)\ne\emptyset$}{
		Derive the solution $k$ to \eqref{eqn-5}\;
		Determine the turning direction $\beta_{k}$ in \eqref{eqn-7}\;
		Choose $\mathbf{a}_{k},\mathbf{b}_{k},\mathbf{c}_{k},\mathbf{d}_{k}$ as in \eqref{eqn-10}-\eqref{eqn-11} and \eqref{eqn-14}-\eqref{eqn-15}\;
		Set $\eta_{k}^{\mathbf{a}},\eta_{k}^{\mathbf{b}},\eta_{k}^{\mathbf{c}},\eta_{k}^{\mathbf{d}}$ as in \eqref{eqn-13} and \eqref{eqn-16}\;
		Derive $\rho_{k}$ and $\delta_{k}$ from \eqref{eqn-18}\;
        Generate $\mathcal{H}_{k}$ in \eqref{eqn-17}\;
		Establish the reference trajectory $\mathcal{L}(t)$ in \eqref{eqn-19}\;
	}	
    \KwOut{$\mathcal{L}(t)$}
\caption{Local Modification}
\label{algo1}
\end{algorithm2e}

Connecting $\mathcal{L}_{k}^{1},\mathcal{C}_{k}^{1},\mathcal{H}_{k},\mathcal{L}_{k}^{2}$ and $\mathcal{C}_{k}^{2}$ in sequence, we obtain the modified reference trajectory
\begin{align}
	\label{eqn-19}
	\mathcal{L}(t)=\mathcal{L}_{k}^{1}\cup\mathcal{C}_{k}^{1}\cup\mathcal{H}_{k}\cup\mathcal{L}_{k}^{2}\cup\mathcal{C}_{k}^{2},
\end{align}
which avoids the obstacle determined via \eqref{eqn-5} while is not collision-free for the other obstacles in $\mathcal{K}(t)$. Hence, at each time, the local modification checking and execution are implemented iteratively until $\mathcal{K}(t)=\emptyset$, which is summarized in Algorithm~\ref{algo1}. In the iterative process, $\mathcal{L}_{k}^{1},\mathcal{C}_{k}^{1},\mathcal{H}_{k},\mathcal{L}_{k}^{2}$ and $\mathcal{C}_{k}^{2}$ are addressed respectively.

The feasibility of Algorithm \ref{algo1} depends on the solvability of \eqref{eqn-5} and \eqref{eqn-18}. Due to $|\mathcal{K}(t)|\le K$, \eqref{eqn-5} can be solved via existing methods like enumeration \citep{Richard1984EC}. In \eqref{eqn-18}, the constraints can be converted into polynomial inequalities, because $\mathcal{E}_{k}$ can be represented by
\begin{align*}
	\mathcal{E}_{k}:=\{\mathfrak{p}\in\mathbb{R}^{2}:e_{i}(\mathfrak{p})\le0,i=1,\ldots,\mathfrak{n}\},
\end{align*}
where $\mathfrak{n}\in\mathbb{N}_{+}$ and $e_{i}:\mathbb{R}^{2}\rightarrow\mathbb{R}$ is polynomial function. Hence, \eqref{eqn-18} is a polynomial optimization problem, whose solvers can be found in \citep{Jean2015Polynomial}. Therefore, Algorithm~\ref{algo1} is feasible. Due to the choice of $\mathbf{a}_{k},\mathbf{b}_{k},\mathbf{c}_{k},\mathbf{d}_{k}$ and the determination of $\rho_{k},\delta_{k}$, the completeness of Algorithm~\ref{algo1} is guaranteed. That is, using Algorithm~\ref{algo1}, a locally collision-free reference trajectory can be generated. To ensure the global collision avoidance, Algorithm \ref{algo1} will be implemented iteratively with the move of the robot.

\section{Adaptive Tracking Control}
\label{sec-ATCS}

Once the reference trajectory $\mathcal{L}(t)$ is generated and ensured to be locally collision-free, the robot will track $\mathcal{L}(t)$ to accomplish its task. In order to track $\mathcal{L}(t)$, we utilize the discretization techniques to develop an adaptive tracking control strategy (ATCS) in this section. 

To begin with, the reference trajectory to be tracked is divided into the following five parts:
\begin{align*}
	\mathcal{T}_{1}(t)&=\mathcal{L}_{k}^{1}\setminus\mathcal{T}_{0}(t),  \\
	\mathcal{T}_{2}(t)&=\mathcal{C}_{k}^{1},  \quad
	\mathcal{T}_{3}(t)=\mathcal{H}_{k},  \quad
	\mathcal{T}_{4}(t)=\mathcal{L}_{k}^{2}, \quad
	\mathcal{T}_{5}(t)=\mathcal{C}_{k}^{2}.
\end{align*}
Each part is of the form
\begin{align*}
    \mathcal{T}_{i}(t):=\{\mathbf{p}\in\mathbb{R}^{2}:\mathbf{x}\in[\underline{\mathbf{x}}_{i},\overline{\mathbf{x}}_{i}],\mathbf{y}=\ell_{i}(\mathbf{x})\},\quad i=\{1,\ldots,5\}.
\end{align*}
Then, $\mathcal{T}_{1}(t),\ldots,\mathcal{T}_{5}(t)$ are discretized along the $\mathbf{x}$-axis such that the tracking task can be transformed into finite sub-tasks of tracking points sequentially. In particular, for $i\in\{1,3,4\}$, $\mathcal{T}_{i}(t)$ is discretized as the points:
\begin{align*}
	&\mathbf{q}_{i}^{0}=(\underline{\mathbf{x}}_{i},\ell_{i}(\underline{\mathbf{x}}_{i})),\ldots,\mathbf{q}_{i}^{l}=\left(\underline{\mathbf{x}}_{i}+l\alpha,\ell_{i}(\underline{\mathbf{x}}_{i}+l\alpha)\right),\ldots, \\
	&\qquad\qquad\qquad\qquad\qquad\qquad\qquad\qquad\quad\mathbf{q}_{i}^{L_{i}}=(\overline{\mathbf{x}}_{i},\ell_{i}(\overline{\mathbf{x}}_{i})),
\end{align*}
and for $i\in\{2,5\}$, $\mathcal{T}_{i}$ is discretized as the points:
\begin{align*}
	&\mathbf{q}_{i}^{0}=(\overline{\mathbf{x}}_{i},\ell_{i}(\overline{\mathbf{x}}_{i})),\ldots,\mathbf{q}_{i}^{l}=\left(\underline{\mathbf{x}}_{i}-l\alpha,\ell_{i}(\underline{\mathbf{x}}_{i}-l\alpha)\right),\ldots, \\ &\qquad\qquad\qquad\qquad\qquad\qquad\qquad\qquad\quad\mathbf{q}_{i}^{L_{i}}=(\underline{\mathbf{x}}_{i},\ell_{i}(\underline{\mathbf{x}}_{i})),
\end{align*} 
where $l\in\{1,\ldots,L_{i}-1\},L_{i}=\lceil\|\overline{\mathbf{x}}_{i}-\underline{\mathbf{x}}_{i}\|/\alpha\rceil$, $\lceil\cdot\rceil$ is the ceil function, and $\alpha>0$ is the discretization parameter. Since the abscissa of the starting point of $\mathcal{T}_{1}(t)$ is $\mathbf{x}(t)$ and $\mathcal{T}_{1}(t),\ldots,\mathcal{T}_{5}(t)$ are connected in sequence, we have
\begin{align*}
    \mathbf{x}(t)=\underline{\mathbf{x}}_{1},\qquad  \mathbf{q}_{i}^{L_{i}}=\mathbf{q}_{i+1}^{0},~i=1,\ldots,4.
\end{align*}
Hence, we set $\mathbf{q}_{1}^{1},\ldots,\mathbf{q}_{1}^{L_{1}},\ldots,\mathbf{q}_{5}^{1},\ldots,\mathbf{q}_{5}^{L_{5}}$ as temporary target positions to transform the tracking task into the sub-tasks of tracking $\sum_{i=1}^{5}L_{i}$ points in order.

Let $\varepsilon\in(0,\alpha)$ be a threshold to verify the tracking performance at each discretized point. If $\|\mathbf{p}(t)-\mathbf{q}_{i}^{l-1}\|\le\varepsilon$, then the robot accomplishes the sub-task of reaching $\mathbf{q}_{i}^{l-1}$ at the time $t$ and needs to accomplish the next sub-task of moving from $\mathbf{p}(t)$ to $\mathbf{q}_{i}^{l}$. To this end, the following optimization problem is formulated.
\begin{align}
    \label{eqn-20}
    \begin{aligned}
        \min\nolimits_{u_{i}^{l}(t)} &~~ \|p(t)-q_{i}^{l}\|^{2}+\mu\|u_{i}^{l}(t)\|^{2} \\
        \text{subject to} &~~ \eqref{eqn-1},~~ \xi_{i}^{l}=\xi(t_{i}^{l-1}),~~ u_{i}^{l}(t)\in\mathbb{U},
    \end{aligned}
\end{align}
where $q_{i}^{l}=\mathbf{R}^{-1}\mathbf{q}_{i}^{l}+s$, $\mu>0$ is a weighted parameter, $t_{i}^{l}>0$ is the time to reach $\mathbf{q}_{i}^{l-1}$ and $\xi_{i}^{l}\in\mathbb{R}^{n}$ is the initial state of this sub-task. The constraint $\xi_{i}^{l}=\xi(t_{i}^{l-1})$ is the connection condition, which ensures both the trajectory smoothness and the motion direction continuity. Since the cost function is quadratic and the constraints are linear, \eqref{eqn-20} can be solved via existing solvers, like AROC \citep{Kochdumper2021aroc}. 

\begin{table*}[!t]
	\caption{Comparisons among different methods in the simplified environment}
	\vspace{-0.2cm}
	\label{table-1}
	\begin{center}
		\begin{tabular}{c|c|c|c|c|c}
			\hline
			& Control Effort & Trajectory Length & Memory Requirement & Planning Time & Total Computation Time\\
			\hline
			RPCS & 227.55 & \textbf{22.99} & \textbf{0.63}GB & \textbf{0.08}s & \textbf{0.19}s \\
			\hline
			LSCM & 233.32 & 23.47 & 2.26GB & 1.37s & 3.07s \\
			\hline
			HPPM & 490.99 & 53.31 & 2.59GB & 8.34s & 10.56s \\
			\hline
			CLF-CBF & 788.44 & 107.91 & 0.73GB & - & 591.37s \\
			\hline
			CILQR & \textbf{117.89} & 24.76 & 1.14GB & - & 4.91s \\
			\hline
			DWA & 204.89 & 23.51 & 13.08GB & - & 3372.07s \\
			\hline
		\end{tabular}
	\end{center}
\end{table*}

When deriving and executing $u_{i}^{l}(t)$, Algorithm \ref{algo1} is executed to ensure the collision avoidance of the remaining points $\mathbf{q}_{i}^{l+1},\ldots,\mathbf{q}_{5}^{L_{5}}$. Once the local modification in Algorithm \ref{algo1} is executed, the corresponding temporary target positions are re-computed. To ensure the collision avoidance, the re-computation of $\mathbf{q}_{i}^{l+1}$ needs to be done before or upon reaching $\mathbf{q}_{i}^{l}$, which requires $\alpha<\gamma$. In the above process, deriving and executing $u_{i}^{l}(t)$ and implementing Algorithm \ref{algo1} are to be done in parallel and iteratively, which is the core of the RPCS. The parallel and iterative execution results in the sub-controllers $u_{1}^{1}(t),\ldots,u_{5}^{L_{5}}(t)\in\mathbb{U}$, which are combined as the following controller
\begin{align}
	\label{eqn-21}
	u(t)=u_{i}^{l}(t),\quad i\in\{1,\ldots,5\},l\in\{1,\ldots,L_{i}\}. 
\end{align} 
With \eqref{eqn-21}, the robot \eqref{eqn-1} can accomplish the original task via a smooth and globally collision-free trajectory.

The performance of the ATCS depends on $\alpha$ and $\varepsilon$. With the decrease of $\alpha$, $\sum_{i=1}^{5}L_{i}$ increases. As $\varepsilon$ decreases, it takes longer time to reach each temporary target position. Hence, the decrease of $\alpha$ and $\varepsilon$ results in better tracking performance and larger computation time. How to choose $\alpha$ and $\varepsilon$ depends on the trade-off between the computation time and the tracking performance.

\begin{figure}[!t] 
	\begin{center}
		\begin{picture}(150,90)
			\put(-45,-11){\resizebox{80mm}{37mm}{\includegraphics[width=2.5in]{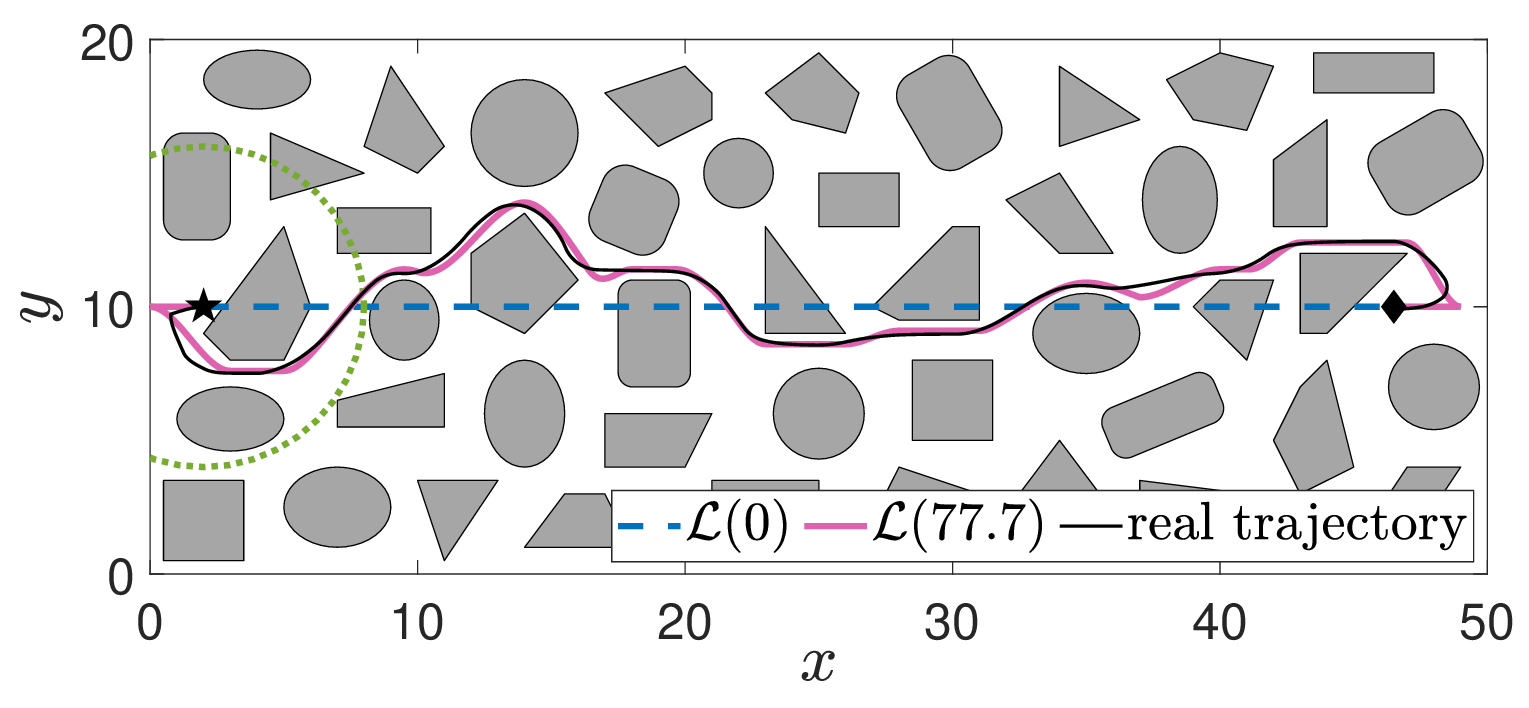}}}
		\end{picture}
	\end{center}
	\caption{Illustration of the reference and real trajectories under the proposed RPCS. The green circle is the sensing boundary at $t=0$. The gray regions are the obstacles enlarged by the robot size $r$.}
	\label{fig-5}
\end{figure}

\begin{figure}[!t] 
	\begin{center}
		\begin{picture}(150, 55)
			\put(-39,-10){\resizebox{80mm}{25mm}{\includegraphics[width=2.5in]{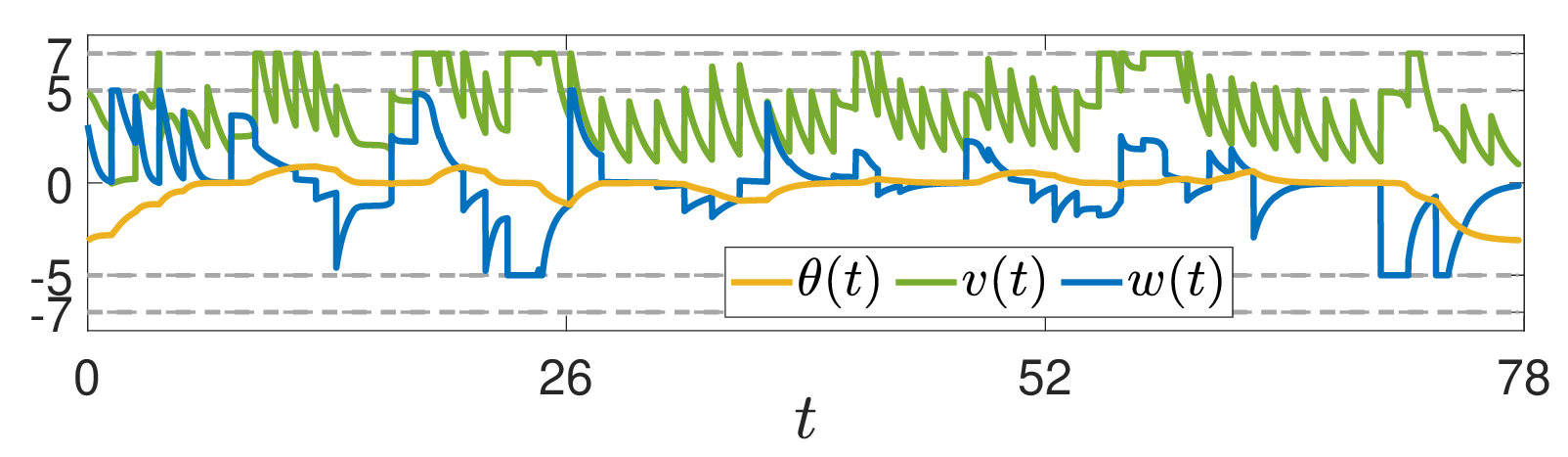}}}
		\end{picture}
	\end{center}
	\caption{Illustration of the motion direction $\theta(t)$ and the control input $(v(t),w(t))$ under the RPCS.}
	\label{fig-6}
\end{figure}

\section{Simulation Results}
\label{sec-simulation}

In this section, the proposed RPCS is first illustrated via a numerical example, and then compared with some existing methods. All computations are executed using Matlab R2024a on a desktop computer with 3.30 GHz Intel Core i3-12100 and 24GB RAM. 
We consider the mobile robot with the radius $r=1$, the sensing radius $\gamma=6$ and the dynamics below:
\begin{align*}
	\begin{bmatrix}
		\dot{p}(t) \\ \dot{\theta}(t)
	\end{bmatrix}=\begin{bmatrix}
		\cos(\theta(t)) & \sin(\theta(t)) & 0 \\ 0 & 0 & 1
	\end{bmatrix}^{\top}\begin{bmatrix}
		v(t) \\ w(t)
	\end{bmatrix}
\end{align*}
where $\xi(t):=(p(t),\theta(t))\in\mathbb{R}^{3}$ is the state and $u(t):=(v(t),w(t))\in\mathbb{U}:=[-7,7]\times[-5,5]$ is the control input. In particular, $p(t):=(x(t),y(t))\in\mathbb{R}^2$ is the position, $\theta(t)\in\mathbb{R}$ is the motion direction, $v(t)\in[-7,7]$ is the linear velocity and $w(t)\in[-5,5]$ is the angular velocity. 

\textbf{Example:} In the obstacle-cluttered environment (see Fig.~\ref{fig-5}), the task of the robot is to move from the initial position $s=(2,10)$ to the target position $g=(46.5,10)$ without the collisions. Let $\rho_{k}=2r,\alpha=0.9$ and $\mu=0.05$. The RPCS is implemented. First, $s$ and $g$ are connected as the initial reference trajectory $\mathcal{L}(0)$; see Fig.~\ref{fig-5}. Then, Algorithm 1 is executed to modify the reference trajectory locally for the collision avoidance, while the ATCS is used to drive the robot to the collision-free temporary target positions. In this way, the robot reaches the target position $g$ at the time $t=77.7$s. The resulting reference trajectory and the real-time trajectory are collision-free, which is given in Fig.~\ref{fig-5} and implies the task accomplishment. From the constraints in \eqref{eqn-20}, the real-time trajectory is smooth, while the reference trajectory is not differentiable at $(0,10)$ and $(49,10)$. The motion direction and control input are presented in Fig.~\ref{fig-6}, where the motion direction is continuous and the control input is bounded.

\begin{figure}[!t] 
	\begin{center}
		\begin{picture}(150,107)
			\put(-38,-13){\resizebox{80mm}{44mm}{\includegraphics[width=2.5in]{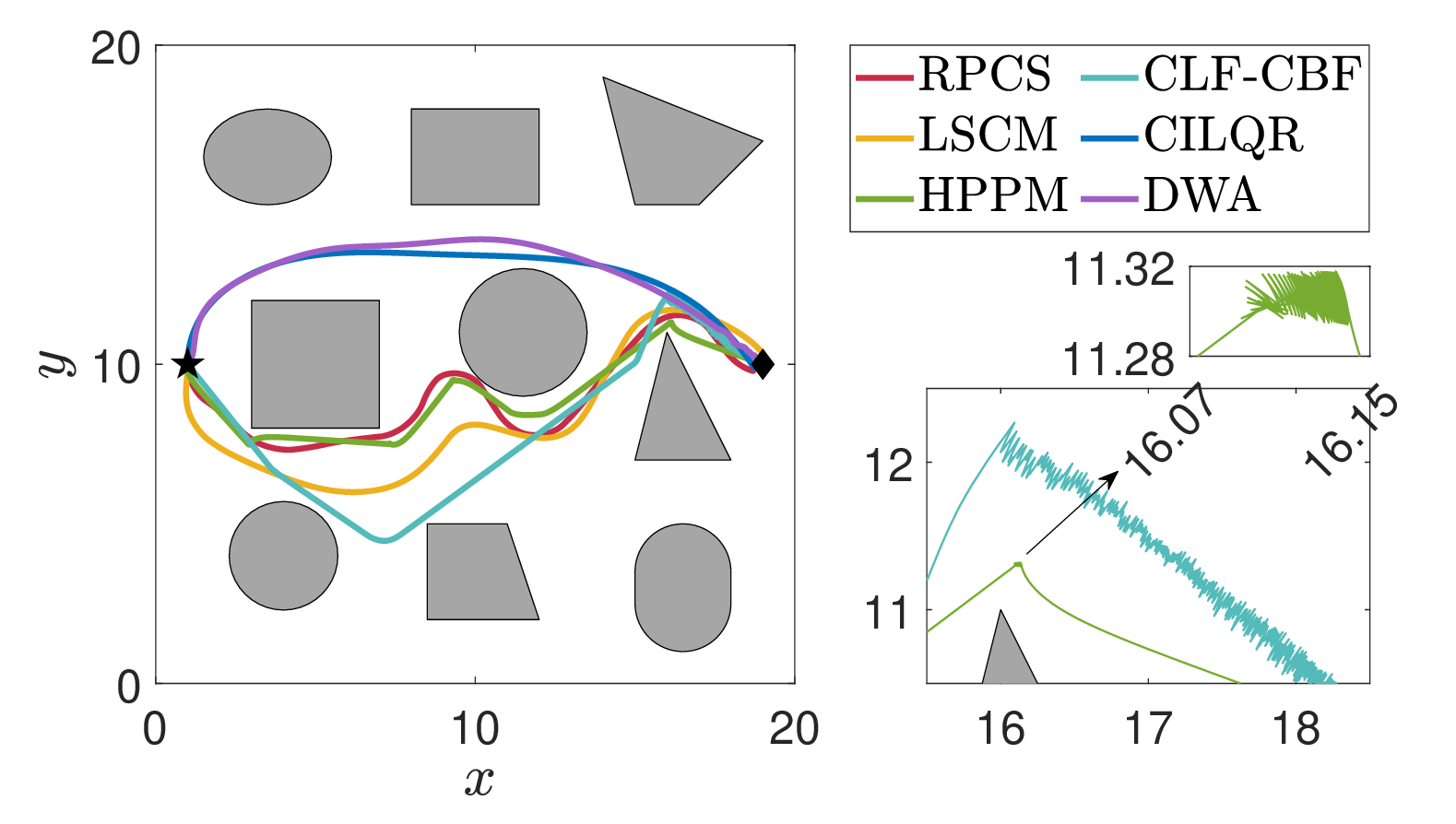}}}
		\end{picture}
	\end{center}
	\caption{Illustration of the trajectories under the RPSC, LSCM, HPPM, CLF-CBF, CILQR and DWA.}
	\label{fig-7}
\end{figure}

\textbf{Comparison:} The proposed RPCS is compared the linear safe corridor based method (LSCM) \citep{Park2022Online}, hybrid path planning based method (HPPM) \citep{Zhong2020HPP}, control Lyapunov function and control barrier function based method (CLF-CBF) \citep{Jankovic2018RobustCBF}, constrained iterative linear quadratic regulator (CILQR) \citep{Chen2019CILQR} and dynamic window approach (DWA) \citep{Fox1997DWA}. The CLF-CBF, CILQR and DWA belong to the first direction of the local approaches, which is to repeatedly implement the global approaches with the change of the partially-known environment information. The LSCM and HPPM belong to the second direction, which consists of the generation, modification and tracking of the reference trajectory. With all the compared methods, the task in the above example fails due to the cluttered and dense obstacles in Fig.~\ref{fig-5}.

For the better comparison, we consider a simplified environment in Fig.~\ref{fig-7}. The initial and target positions are $s=(2,10)$ and $g=(8,10)$, respectively. Under the RPCS, LSCM, HPPM, CLF-CBF, CILQR and DWA, the task can be accomplished. The position trajectories are shown in Fig.~\ref{fig-7}. The performance indicators including the control effort, trajectory length, memory requirement, planning time and total computation time are shown in Table~\ref{table-1}. From Fig.~\ref{fig-7} and Table~\ref{table-1}, the proposed RPCS generates a smooth and shortest trajectory to accomplish the task with the minimal memory requirement, planning time and total computation time. The CILQR has the minimal control effort, but relies on the initial guess of the control input. The inaccurate or wrong initial guess may result in the task failure. The large trajectory lengths under the HPPM and CLF-CBF mainly result from the jitter phenomena; see Fig.~\ref{fig-7}. 

\section{Conclusion}
\label{sec-conclusion}

This paper studied the motion control problem for mobile robots with partial information of the obstacle-cluttered environments. We proposed the RPCS such that the robot moved from an initial position to a target position without the collisions via reactive planning and adaptive tracking control. Simulation results were provided to show the efficacy and advantages of the proposed RPCS. Future work will extend the RPCS to multi-robot case.

\bibliography{ref_original}
\end{document}